\documentclass[runningheads]{llncs}
\usepackage{graphicx}
\usepackage{setspace}  
\usepackage{booktabs}  
\usepackage{adjustbox}  
\usepackage{multirow}  
\usepackage{xcolor}  
\usepackage{amsmath}  
\usepackage{amssymb}  
\usepackage{cite}  

\begin{document}
\title{Simple Techniques for Enhancing Sentence Embeddings in Generative Language Models}
%
%
\author{Bowen Zhang \and
Kehua Chang \and
Chunping Li}
\authorrunning{Bowen et al.}
\titlerunning{Enhancing Sentence Embeddings in Generative Language Models}
%
\institute{
School of Software, Tsinghua University\\
\email{zbw23@mails.tsinghua.edu.cn}\\
\email{changkehua@gmail.com}\\
\email{cli@tsinghua.edu.cn}
}
\maketitle              
\begin{abstract}
Sentence Embedding stands as a fundamental task within the realm of Natural Language Processing, finding extensive application in search engines, expert systems, and question-and-answer platforms. With the continuous evolution of large language models such as LLaMA and Mistral, research on sentence embedding has recently achieved notable breakthroughs. However, these advancements mainly pertain to fine-tuning scenarios, leaving explorations into computationally efficient direct inference methods for sentence representation in a nascent stage. This paper endeavors to bridge this research gap. Through comprehensive experimentation, we challenge the widely held belief in the necessity of an Explicit One-word Limitation for deriving sentence embeddings from Pre-trained Language Models (PLMs). We demonstrate that this approach, while beneficial for generative models under direct inference scenario, is not imperative for discriminative models or the fine-tuning of generative PLMs. This discovery sheds new light on the design of manual templates in future studies. Building upon this insight, we propose two innovative prompt engineering techniques capable of further enhancing the expressive power of PLMs' raw embeddings: Pretended Chain of Thought and Knowledge Enhancement. We confirm their effectiveness across various PLM types and provide a detailed exploration of the underlying factors contributing to their success. Our code has been made publicly available at \url{https://github.com/ZBWpro/PretCoTandKE}.
 
\end{abstract}

\section{Introduction}

The advent of generative Pre-trained Language Models (PLMs), exemplified by GPT \cite{GPT3-NIPS-2020}, OPT \cite{OPT-2022}, and LLaMA \cite{LLaMA-2023}, marks a significant milestone in the field of Natural Language Processing (NLP). Boasting immense parameter sizes and extensive pre-training corpora, these models have demonstrated unparalleled proficiency in multi-task processing and zero-shot reasoning. Consequently, there is a growing interest in directly leveraging these PLMs for a wide array of downstream tasks, including the derivation of sentence embeddings.

Sentence embeddings are high-dimensional vector representations that encapsulate the semantic essence of original sentences. These embeddings are pivotal not only as features for neural network tackling complex tasks but also in applications such as information retrieval, text clustering, recommendation systems, topic modeling, and retrieval-augmented generation. Thus, their research and practical implications are substantial.

Over the past few years, research in this direction has predominantly revolved around discriminative models like BERT \cite{BERT-NAACL-2019} and RoBERTa \cite{RoBERTa-2019}. Nonetheless, the semantic space anisotropy inherent in these models \cite{BERT-flow-EMNLP-2020} often mandates fine-tuning with considerable datasets to produce high-quality sentence embeddings. Among various fine-tuning approaches, contrastive learning, which aims to bring similar samples closer while distancing disparate ones, has proven to be the most effective. However, to prevent model collapse and ensure sufficient reference items for sentence vector calculation, contrastive learning methods necessitate large training batch sizes, thereby consuming massive computational resources.

In response to this challenge, some cutting-edge efforts have shifted their focus to the recently developed generative models, hoping to leverage their advanced text comprehension abilities to directly encode input sentences without additional backward propagation. Given the disparity between sentence representation and auto-regressive language modeling, this endeavor demands sophisticatedly designed prompts. The pioneering work, PromptEOL \cite{PromptEOL-2023}, find that by introducing an Explicit One word Limitation (EOL) into the prompt and constructing a template akin to `This sentence: ``[X]" means in one word: ``', it can significantly enhance the embedding obtained from generative models. Subsequent studies \cite{Angle-2023} \cite{DeeLM-2023} have followed this practice while also combining EOL with contrastive learning for fine-tuning Large Language Models (LLMs). Nevertheless, even with efficient fine-tuning strategies like QLoRA \cite{QLoRA-NIPS-2024}, training a neural network with a 7B-scale LLM as the backbone still consumes more GPU memory than fully fine-tuning a 110M-scale BERT. Moreover, we elucidate in subsequent sections that EOL chiefly benefits raw embeddings from generative models and is not requisite for discriminative models or fine-tuning generative models.

Therefore, to reconcile the demand for high-quality sentence embeddings with the need for computational resource conservation, it is advisable to adopt PLMs with relatively high parameter scale while avoiding gradient updates. Given that the quality of embeddings directly encoded from input sentences is generally low \cite{PromptEOL-2023}, this paper designs two simple yet effective prompt engineering methods: Pretended Chain of Thought (CoT) and Knowledge Enhancement, to ameliorate the sentence representation capabilities of generative models under direct inference settings. Both techniques involve appending a fixed prefix before the EOL to harness PLMs' contextual learning ability to its fullest potential.
 
Specifically, Pretended CoT, inspired by the design of Zero-shot CoT \cite{CoT-NIPS-2022}, mainly serves an ``inspirational" role since we do not need the model to output intermediate reasoning steps. Knowledge Enhancement, on the other hand, provides explicit guidance to the model by conveying human experience in text summarization through prompts. Our comprehensive experimental outcomes affirm that Pretended CoT and Knowledge Enhancement not only optimize memory efficiency but also significantly elevate the performance of raw embeddings from generative models across multiple Semantic Textual Similarity (STS) benchmarks.

The main contributions of this paper are outlined as follows:
\begin{itemize}
\item Our investigations reveal that the widely applied Explicit One-word Limitation (EOL) is actually a sentence embedding derivation technique for direct inference scenarios with generative PLMs. It does not always yield optimal performance in discriminative models or fine-tuning generative models. This finding opens new avenues for future research on manual template design.
\item This paper proposes two concise and potent prompt engineering methods, Pretended CoT and Knowledge Enhancement, that markedly enhance the quality of embeddings produced by generative models on top of the original prompt.
\item Through rigorous experimentation with PLMs of varying types and sizes, we validate the superiority of our proposed techniques and offer an in-depth analysis of their success. In the spirit of fostering reproducibility and inspiring subsequent research, we have made our code publicly accessible.
\end{itemize}

\section{Related Work}

The exploration of sentence embeddings with PLMs constitutes a vibrant area of research within Natural Language Processing and Computational Linguistics. This body of work diverges into three primary streams: supervised fine-tuning, unsupervised fine-tuning, and direct inference. Each paradigm offers unique insights while presenting distinct challenges, as delineated below.

\textbf{Supervised Fine-tuning}: This strand of research leverages high-quality annotated corpora to enhance models' encoding capabilities. A landmark in this domain, Sentence-BERT \cite{Sentence-BERT-EMNLP-2019} designs a Siamese neural network architecture with shared parameters and updates model via the NLI dataset \cite{snli, mnli}. Following this, SimCSE \cite{SimCSE-EMNLP-2021} filters the NLI dataset and organizes it into triplets to adjust the sentence vector distribution of BERT through contrastive learning. Further studies like PromptEOL \cite{PromptEOL-2023}, AnglE \cite{Angle-2023}, and DeeLM \cite{DeeLM-2023} extend this methodology to generative PLMs such as OPT \cite{OPT-2022}, LLaMA \cite{LLaMA-2023}, and LLaMA2 \cite{LLaMA2-2023}. While supervised learning methods consistently outperform across a spectrum of benchmarks, their dependency on labeled datasets limits their applicability in a broader context.

\textbf{Unsupervised Fine-tuning}: Focusing on the potential of models trained with unlabelled corpora, unsupervised fine-tuning has emerged as a fertile ground for innovation. This realm is particularly enriched by contrastive learning methods and data augmentation techniques. ESimCSE \cite{ESimCSE-COLING-2022}, for example, advances SimCSE by introducing random word repetition in the input text to construct positives for original sentences. PromptBERT \cite{PromptBERT-EMNLP-2022}, in contrast, utilizes different manual templates to create positive instances. Additionally, SSCL \cite{SSCL-ACL-2023} demonstrates the efficacy of using intermediate layer outputs from PLMs as hard negatives in the InfoNCE Loss \cite{InfoNCELoss-2018} framework. CoT-BERT \cite{CoT-BERT-2023}, representing the current state-of-the-art, nearly matches the performance of supervised approaches across multiple STS benchmarks by relying solely on the intrinsic capabilities of PLMs. Despite these encouraging advancements, unsupervised fine-tuning methods do not inherently present advantages in memory efficiency, given their reliance on large batch sizes to facilitate effective contrastive learning.

\textbf{Direct Inference}: Efficient sentence representation methods that directly utilize raw embeddings of PLMs without parameter updates are increasingly attracting attention in the NLP community. Among these, PromptBERT \cite{PromptBERT-EMNLP-2022} showcases that meticulously designed templates tailored to mimic the Masked Language Modeling (MLM) task can exploit BERT’s original output vectors for commendable performance. Building on this insight, PromptEOL \cite{PromptEOL-2023} targets generative PLMs and advocates explicitly prompting the model to condense the original sentence into a single word. This strategy markedly taps into the latent potential of generative models and surpasses the direct inference results of PromptBERT. Considering the multifaceted nature of refining textual semantics, MetaEOL \cite{MetaEOL-2024} employs ChatGPT-4 to craft diverse templates, feeding them into the PLM and extracting the sentence representation through average pooling. Although the rationale behind this method is theoretically sound, encoding the same sentence multiple times with distinct templates diverges from the original intention of offline computation.

\section{Methodology}

In this chapter, we first demonstrate through pilot experiments in Sections~\ref{sec:eol_dis} and \ref{sec:eol_gen} that the prevailing paradigm in current sentence representation research, Explicit One-word Limitation (EOL), primarily functions to enhance the direct inference effects of generative PLMs. It does not consistently yield superior outcomes in discriminative models or when fine-tuning generative PLMs. Therefore, explorations surrounding EOL should focus more on scenarios that directly apply PLM raw embeddings for inference. Subsequently, in Section~\ref{sec:prompt}, we provide detailed descriptions of two innovative prompt engineering techniques proposed in this paper, Pretended Chain of Thought (CoT) and Knowledge Enhancement, for further elevating the semantic expressiveness of PLMs' raw embeddings.

\subsection{EOL in the Context of Discriminative Models}
\label{sec:eol_dis}

The design philosophy of EOL stems from an analysis of inference outcomes when employing an unmodified BERT model on the STS-B development set \cite{STS-B}. This analysis contrasts two highly similar manual templates, distinguished solely by the presence or absence of a terminal period:

\centerline{\texttt{This sentence : "[X]" means [MASK] \textcolor{red}{.}}}
\centerline{\texttt{This sentence : "[X]" means [MASK]}}

Researchers of PromptEOL discover that when utilizing BERT$_{\text{base}}$ and taking the output vector corresponding to the [MASK] token as the sentence embedding for the original sentence [X], the template terminating with a period demonstrates markedly superior performance compared to its counterpart without the period \cite{PromptEOL-2023}. This observation leads to the hypothesis that the period at the end of the template plays a crucial role in directing the model to condense the text's semantics into a single word, thereby yielding superior results. This insight forms the basis for applying a similar ``in one word” restriction in generative model prompts.

Expanding upon this premise, our study embarks on a more granular investigation. We commence with a series of comparative experiments centered on BERT$_{\text{base}}$, examining the impact of the number of [MASK] tokens and the selection of terminal characters in templates. Adopting the identical experimental setup as PromptEOL, we utilize `This sentence: ``[X]" means [MASK].' as our baseline template and initialize BERT from its original pre-trained state with its performance measured on the STS-B development set via Spearman correlation coefficient. The corresponding results are presented in Table~\ref{tab:bert_dev}.
\begin{table*}[htbp]
\caption{Performance of BERT$_{\text{base}}$ on the STS-B development set under varying manual template configurations, with the last column indicating the Spearman correlation coefficient (multiplied by 100).}
\centering
\small
    \begin{tabular}{cccc}
    \toprule
    \bf{[MASK]} & \bf{EOS} & \bf{STS-B dev} \\
    \midrule
    1 & None & 27.21 \\
    1 & [SEP] & 32.54 \\
    \midrule
    1 & $.$ & 73.61 \\
    2 & $.$ & 73.94 \\
    3 & $.$ & \bf 74.50 \\
    4 & $.$ & 71.83 \\ 
    \midrule
    1 & $!$ & \bf 77.59 \\
    2 & $!$ & 76.55 \\
    3 & $!$ & 75.99 \\
    4 & $!$ & 73.45 \\
    \midrule
    1 & $?$ & 75.32 \\
    2 & $?$ & 75.19 \\
    3 & $?$ & \bf 75.37 \\
    4 & $?$ & 73.62 \\
    \bottomrule
    \end{tabular}
\label{tab:bert_dev}
\end{table*}

In Table~\ref{tab:bert_dev}, the column labeled ``[MASK]" denotes the number of [MASK] tokens within the prompt, while the ``EOS" (End of Sentence) column specifies the selected terminal punctuation of the template. For example, a configuration with [MASK] set to 2 and EOS as `!’ results in the prompt `This sentence: ``[X]" means [MASK][MASK] !'. In cases where multiple [MASK] tokens are employed, we calculate the mean of their output vectors as the corresponding sentence representation.

The analysis reveals that omitting a terminal character in the manual template, BERT achieves a mere Spearman correlation coefficient score of 27.21 on the STS-B development set. In contrast, the introduction of a period, exclamation mark, or question mark substantially bolsters the representational capacity of BERT’s raw embeddings, elevating its score to reach over 70. Interestingly, the period emerges as the least effective among these punctuation marks. Additionally, the findings indicate that utilizing a single [MASK] token is suboptimal when the sentence concludes with a period or question mark. In other words, the EOL hypothesis of summarizing the original text into one word does not always hold true for discriminative models like BERT. 

\subsection{EOL in the Context of Generative Models}
\label{sec:eol_gen}

With the establishment that EOL is not universally applicable to discriminative PLMs, we proceed to conduct experiments with generative models such as OPT, LLaMA, LLaMA2, and Mistral \cite{Mistral-2023}. In this endeavor, we craft two novel templates, PromptSTH and PromptSUM, intentionally devoid of the explicit ``in one word" constraint:

\centerline{{PromptEOL} $\rightarrow$ \texttt{This sentence : "[X]" means in one word:\textcolor{red}{"}}}
\centerline{{PromptSTH} $\rightarrow$ \texttt{This sentence : "[X]" means \textcolor{red}{something}}}
\centerline{{PromptSUM} $\rightarrow$ \texttt{This sentence : "[X]" can be summarized \textcolor{red}{as}}}

Mirroring PromptEOL, we derive the sentence embedding for the original sentence [X] from the output vector corresponding to these templates' concluding token, sourced from the PLMs' last hidden layer. These prompts undergo evaluation under two distinct conditions: parameter-efficient fine-tuning and direct inference. Their performance is assessed across seven established STS benchmarks \cite{STS12, STS13, STS14, STS15, STS16, STS-B, SICK-R} to ascertain their average Spearman correlation coefficient, as illustrated in Table~\ref{tab:gen_eol}.
\begin{table*}[ht]
\caption{Average Spearman correlation coefficient (multiplied by 100) across seven STS benchmarks using various manual templates, under both parameter-efficient fine-tuning and direct inference modalities.}
\centering
    \begin{tabular}{ccc}
    \toprule
    \bf{Method} & \bf{Fine-tuning} & \quad \bf{Direct Inference} \\      
    \midrule
    \multicolumn{3}{c}{\it{Implementation on $OPT_{\rm 6.7b}$}} \\
    \multicolumn{1}{l}{PromptEOL} & 85.52 & \bf 72.10 \\
    \multicolumn{1}{l}{PromptSTH} & 85.51 & 56.82 \\
    \multicolumn{1}{l}{PromptSUM} & \bf 85.57 & 68.40 \\
    \midrule \midrule
    \multicolumn{3}{c}{\it{Implementation on $LLaMA_{\rm 7b}$}} \\
    \multicolumn{1}{l}{PromptEOL} & \bf 85.48 & \bf 68.76 \\
    \multicolumn{1}{l}{PromptSTH} & 85.40 & 52.48\\
    \multicolumn{1}{l}{PromptSUM} & 85.47 & 53.78 \\
    \midrule \midrule
    \multicolumn{3}{c}{\it{Implementation on $LLaMA2_{\rm 7b}$}} \\
    \multicolumn{1}{l}{PromptEOL} & 85.40 & \bf 70.03 \\
    \multicolumn{1}{l}{PromptSTH} & 85.31 & 55.08 \\
    \multicolumn{1}{l}{PromptSUM} & \bf 85.53 & 46.61 \\
    \midrule \midrule
    \multicolumn{3}{c}{\it{Implementation on $Mistral_{\rm 7b}$}} \\
    \multicolumn{1}{l}{PromptEOL} & 85.50 & \bf 73.32 \\
    \multicolumn{1}{l}{PromptSTH} & 85.66 & 58.94 \\
    \multicolumn{1}{l}{PromptSUM} & \bf 85.83 & 61.91 \\
    \bottomrule
    \end{tabular}
\label{tab:gen_eol}
\end{table*}

In Table~\ref{tab:gen_eol}, the ``Fine-tuning" column corresponds to the experimental results obtained by employing the NLI dataset as training corpus, coupled with the application of contrastive learning and QLoRA for fine-tuning. Conversely, the ``Direct Inference" column represents the Spearman correlation scores derived from leveraging the raw embeddings of PLMs. According to the experimental outcomes, the performance disparities among different templates post parameter updates are actually quite minimal. It could even be said that PromptSUM performs better than PromptEOL when fine-tuned. However, within direct inference scenarios, PromptEOL demonstrates a clear advantage. For instance, employing Mistral$_{\rm 7b}$, PromptSUM attains an average Spearman correlation of 85.83 following supervised fine-tuning, slightly higher than PromptEOL's 85.50. Nonetheless, in direct inference, PromptEOL significantly outperforms PromptSUM by 11.41 points (73.32 vs. 61.91).

In conclusion, EOL emerges as a viable technique for augmenting the quality of raw embeddings from generative PLMs. Its utility is less pronounced within discriminative models or amidst the fine-tuning of generative models. Therefore, future discourse on EOL should concentrate on its applicability to direct inference scenarios.

\subsection{Pretended CoT and Knowledge Enhancement}
\label{sec:prompt}

The experimental findings detailed in Table~\ref{tab:gen_eol} demonstrate that while EOL notably enhances the semantic expressiveness of generative PLMs, the direct inference capabilities of 7B-scale models remain inferior to those of unsupervised fine-tuned SimCSE-BERT$_{\text{base}}$, which achieves an average Spearman correlation score of 76.25 across seven STS benchmarks \cite{SimCSE-EMNLP-2021}. This score markedly surpasses the 70.03 performance metric of LLaMA2$_{\rm 7b}$-EOL.

For further improvement, PromptEOL suggests adopting a one-shot format to provide the model with a clear example, so as to facilitate the text summarization process implied by EOL more effectively. While this method adeptly exploits auto-regressive models' contextual learning capacity, its reliance on specific examples limits its universality across PLMs of diverse types and sizes. Each example within the demonstration set must undergo evaluation against a validation set to identify the most suitable choice, a procedure that is both resource-intensive and time-consuming. Alternatively, MetaEOL introduces a strategy involving the definition of meta-tasks, with ChatGPT generating corresponding templates for each. These templates are then amalgamated with the original sentence and processed through the PLM to derive multiple embeddings, thereby enabling the model to analyze various semantic aspects. This method, however, requires repeated encoding of sentences, diminishing efficiency.

This study aims to refine the quality of generative PLM raw embeddings in a more streamline and concise manner. To this end, we propose two plug-and-play prompt engineering techniques: Pretended CoT and Knowledge Enhancement. Their integration with EOL is illustrated in Table~\ref{tab:templates}. 
\begin{table}[htbp]
\caption{Incorporation of Pretended Chain of Thought and Knowledge Enhancement into the EOL Template.}
\centering
\begin{spacing}{1.5}
\begin{tabular}{c}
\toprule
{\bf Explicit One word Limitation}  \\
This sentence : ``[X]" means in one word:``\\
\midrule 
{\bf Pretended Chain of Thought} \\
After thinking step by step , this sentence : ``[X]" means in one word:``\\
\midrule 
{\bf Knowledge Enhancement} \\
The essence of a sentence is often captured by its main subjects and actions,\\
while descriptive terms provide additional but less central details. \\
With this in mind , this sentence : ``[X]" means in one word:``\\ 
\bottomrule 
\end{tabular}
\end{spacing}
\label{tab:templates}
\end{table}

Pretended CoT draws inspiration from Zero-shot CoT, which guides dialogue models to sequentially solve complex reasoning problems by appending ``Let's think step by step." at the end of the prompt. Empirical evidence suggests that this strategy effectively unlocks the potential of large-scale PLMs, enabling them to complete target tasks with higher accuracy. In pursuit of computational resources and efficiency, we choose not to use as large generative models as Zero-shot CoT. In fact, the majority of experiments in this paper are conducted on PLMs at or below 7B scale. Although these models possess some text continuation capabilities, the quality of the generated content is generally low. Therefore, rather than eliciting intermediate reasoning steps, we encourage a meticulous approach to sentence representation by prefacing EOL with ``After thinking step by step,".

Knowledge Enhancement, in contrast, seeks to explicitly infuse the model with human insights into text summarization through tailored prompts. This strategy is designed to guide the model in distilling the original sentence into a single word by adhering to general summarization principles, thereby allocating its attention in a more rational and interpretable manner. Specifically, for brief sentences, we generally consider the subjects and actions bear greater semantic weight, while auxiliary elements are deemed to have a relatively smaller impact. By adding this prior knowledge ahead of EOL, we form the Knowledge Enhanced EOL adopted in this paper.

It is important to highlight that, to minimize operational complexities, the implementations of Pretended CoT and Knowledge Enhancement for all PLMs, regardless of their specific types and sizes, are completely consistent in our research. In forthcoming sections, we will demonstrate through specific experimental results that this unified configuration is sufficient to provide a stable performance improvement for models.

\section{Experiments}
\label{sec:experiment}

This chapter presents the experimental results of our two novel prompt engineering strategies: Pretended CoT and Knowledge Enhancement. Initially, Section~\ref{sec:7b_performance} assesses the efficacy of these methods when integrated with Explicit One-word Limitation (EOL). This evaluation is conducted on prevalent 7B-scale generative PLMs, including OPT$_{\rm 6.7b}$, LLaMA$_{\rm 7b}$, LLaMA2$_{\rm 7b}$, and Mistral$_{\rm 7b}$. Following this, Section~\ref{sec:gpu_memory} provides a comparative analysis of GPU memory consumption between fine-tuning and direct inference approaches. Finally, Section~\ref{sec:scale_performance} examines the impact of PLM parameter sizes on the performance of these techniques, utilizing OPT models ranging from 350M to 13B parameters. 

Regarding the experimental setup, we follow the established practices from prior research, assessing model performance across seven STS benchmarks via the SentEval \cite{SentEval-LREC-2018} toolkit, with the Spearman correlation coefficient score as the core evaluation metric.

\subsection{Efficacy on 7B-Scale Generative PLMs}
\label{sec:7b_performance}

Since our study primarily targets direct inference scenarios, we uniformly load the initial checkpoint of PLMs, eschewing any further parameter updates. In all experimental groups, the configurations of PromptEOL, Pretended CoT, and Knowledge Enhancement remain consistent with Table~\ref{tab:templates}.

For PromptEOL, we adhere to the methodology proposed in the original paper, taking the output vector corresponding to the concluding quotation mark from the model's last hidden layer as the sentence embedding. For Pretended CoT and Knowledge Enhancement, we opt for the encoding from the penultimate hidden layer for the terminal token, as this approach demonstrates superior outcomes. It is noteworthy, however, that even when selecting the same hidden layer, Pretended CoT and Knowledge Enhancement can still exhibit a clear facilitative effect.

Table~\ref{tab:main_results} provides a comprehensive summary of the experimental findings. Notably, Pretended CoT and Knowledge Enhancement significantly amplify the models' performances based on PromptEOL. This improvement is particularly remarkable within the contexts of LLaMA and LLaMA2, where the average Spearman correlation coefficient experiences an impressive uplift of 10.56\% and 10.15\%, respectively. Given that these enhancements are attained merely by appending a fixed prefix to the input template, without the necessity for additional training data or external modifications, the effectiveness of these strategies is manifestly apparent. Furthermore, across all evaluated PLMs, the peak performance on the seven STS tasks is consistently realized through either Pretended CoT or Knowledge Enhancement, affirming the robustness and versatility of our approach.

In addition, we compare our method against three unsupervised fine-tuning approaches based on BERT model: BERT-flow, BERT-whitening \cite{BERT-whitening-2021}, and SimCSE. By employing Knowledge Enhancement to refine LLaMA2$_{\rm 7b}$'s raw embeddings, our strategy attains an average Spearman correlation score of 77.14 without any training, surpassing the results of the aforementioned unsupervised schemes. 
\begin{table*}[ht]
\caption{Spearman’s correlation scores across seven STS benchmarks for various models. $\dagger$: results from SimCSE \cite{SimCSE-EMNLP-2021}. $\ddagger$: results from PromptBERT \cite{PromptBERT-EMNLP-2022}. $\S$: results from Sentence-T5 \cite{Sentence-T5-ACL-2022}.}
\centering
\resizebox{1.0\linewidth}{!}{
    \begin{tabular}{ccccccccc}
    \toprule 
    \bf{Methods} & \bf{STS-12} & \bf{STS-13} & \bf{STS-14} & \bf{STS-15} & \bf{STS-16} & \bf{STS-B} & \bf{SICK-R} & \bf{Avg.} \\
    \midrule \midrule
    \multicolumn{9}{c}{\textbf{Fine-tuned on Unsupervised Datasets}} \\
    \midrule \midrule
    BERT-flow$^{\dagger}_{\text{base}}$ & 58.40 & 67.10 &	60.85 &	75.16 &	71.22 &	68.66 &	64.47 &	66.55 \\
    BERT-whitening$^{\dagger}_{\text{base}}$ & 57.83 & 66.90 & 60.90 & 75.08 & 71.31 & 68.24 & 63.73 & 66.28 \\
    SimCSE-BERT$^{\dagger}_{\text{base}}$ & 68.40 & 82.41 & 74.38 & 80.91 & 78.56 & 76.85 & 72.23 & 76.25 \\
    SimCSE-RoBERTa$^{\dagger}_{\text{base}}$ & 70.16 &  81.77 & 73.24 & 81.36 & 80.65 & 80.22 & 68.56 & 76.57 \\
    \midrule \midrule
    \multicolumn{9}{c}{\textbf{Direct Inference}} \\
    \midrule \midrule
    BERT avg.\(^\ddagger\) & 30.87 & 59.89 & 47.73 & 60.29 & 63.73 & 47.29 & 58.22 & 52.57 \\
    BERT prompt\(^\ddagger\) & 60.96 & 73.83 & 62.18 & 71.54 & 68.68 & 70.60 & 67.16 & 67.85 \\
    ST5-Enc\(^\S\) & 34.97 & 60.19 & 47.59 & 66.40 & 70.62 & 62.83 & 63.57 & 58.02 \\
    \midrule
    \multicolumn{9}{c}{\it{Implementation on $OPT_{\rm 6.7b}$}} \\
    PromptEOL & 60.91 & 80.05 & 67.65 & 75.49 & 80.11 &  72.91 & 67.57 & 72.10 \\
    Pretended CoT & \bf 64.70 & \bf 82.41 & \bf 71.99 & \bf 78.63 & \bf 81.62 & \bf 76.30 & 68.72 & \bf 74.91 \\
    Knowledge Enhancement & 64.30 & 82.13 & 71.35 & 78.48 & 79.74 & 74.83 & \bf 71.34 & 74.60 \\
    \midrule
    \multicolumn{9}{c}{\it{Implementation on $Mistral_{\rm 7b}$}} \\
    PromptEOL & 63.08 & 78.58 &  69.40 & 77.92 & 79.01 &  75.77 & 69.47 & 73.32 \\
    Pretended CoT & \bf 66.45 & \bf 82.04 & \bf 72.24 & \bf 77.93 & \bf 79.36 & \bf 76.66 & 71.06 & \bf 75.11 \\
    Knowledge Enhancement & 60.33 & 81.52 & 71.73 & 77.53 & 77.99 & 74.09 & \bf 74.02 & 73.89 \\
    \midrule
    \multicolumn{9}{c}{\it{Implementation on $LLaMA_{\rm 7b}$}} \\
    PromptEOL & 60.15 & 76.14 & 65.22 & 74.02 & 74.19 & 67.93 & 63.64 & 68.76 \\
    Pretended CoT & \bf 67.36 & \bf 82.02 & \bf 73.68 & \bf 80.88 & \bf 80.69 & \bf 77.47 & 70.01 & \bf 76.02 \\
    Knowledge Enhancement & 63.16 & 81.38 & 71.83 & 80.51 & 78.26 & 77.30 & \bf 73.02 & 75.07 \\
    \midrule
    \multicolumn{9}{c}{\it{Implementation on $LLaMA2_{\rm 7b}$}} \\
    PromptEOL & 58.81 & 77.01 & 66.34 & 73.22 & 73.56 & 71.66 & 69.64 & 70.03 \\
    Pretended CoT & \bf 67.45 & \bf 83.89 & 74.14 & 79.47 & \bf 80.76 & 78.95 & 73.33 & 76.86 \\
    Knowledge Enhancement & 65.60 & 82.82 & \bf 74.48 & \bf 80.75 & 80.13 & \bf 80.34 & \bf 75.89 & \bf 77.14 \\
    \bottomrule
    \end{tabular}%
}
\label{tab:main_results}
\end{table*}

\subsection{Computational Cost Comparison}
\label{sec:gpu_memory}

This subsection quantifies the computational efficiency of direct inference methods utilizing raw embeddings from PLMs, in comparison to contrastive learning approaches that necessitate fine-tuning on extensive datasets. We document the GPU memory consumption for both approaches, with the results detailed in Table~\ref{tab:memory}.
\begin{table}[htbp]
\caption{Comparison of GPU memory consumption across various sentence embedding derivation methods.}
\centering
\begin{adjustbox}{width=\linewidth, center}
\begin{tabular}{ccccc}
    \toprule
    \bf Methods & \bf PLMs & \bf Params & \bf Settings & \bf Memory Usage \\
    \midrule
    SimCSE & BERT & 110M & Fine-tuning (full) & 58 GB \\
    PromptEOL & LLaMA2 & 7B & Fine-tuning (QLoRA) & 93 GB \\
    Pretended CoT & LLaMA2 & 7B & Direct Inference & \bf{36} GB \\
    Knowledge Enhancement & LLaMA2 & 7B & Direct Inference & \bf{39} GB \\
    \bottomrule
\end{tabular}
\end{adjustbox}
\label{tab:memory}
\end{table}

The results indicate that parameter updates on PLMs, particularly at the 7B-scale, require substantial computational power. In stark contrast, our Pretended CoT and Knowledge Enhancement strategies ingeniously leverage the inherent capabilities of generative PLMs without the need for any gradient updates, thereby reducing GPU memory usage to under 40GB. Given the exceptional performance demonstrated by Pretended CoT and Knowledge Enhancement in Section~\ref{sec:7b_performance}, they hold significant potential for practical implementations.

\subsection{PLM Performance across Different Scales}
\label{sec:scale_performance}

To examine the efficacy of Pretended CoT and Knowledge Enhancement across PLMs of various sizes, we utilize the same experimental setup as detailed in Section~\ref{sec:7b_performance}, assessing the performance of these techniques on OPT models with parameters ranging from 350M to 13B. The results of these evaluations are encapsulated in Table~\ref{tab:opt_sts}, wherein the ``Spearman" column denotes the average Spearman correlation coefficient score achieved by models across seven STS benchmarks.
\begin{table}[htbp]
\caption{Performance of Pretended CoT and Knowledge Enhancement across varying scales of OPT models.}
\centering
\begin{tabular}{ccc}
    \toprule
    \bf PLMs & \bf Methods  & \bf Spearman\\
    \midrule
    \multirow{3}{*}{OPT$-{\rm 350m}$} 
    & PromptEOL & 65.03 \\
    & Pretended CoT & \bf 66.16 \\
    & Knowledge Enhancement & 64.18 \\
    \midrule
    \multirow{3}{*}{OPT$-{\rm 1.3b}$} 
    & PromptEOL & 73.18 \\
    & Pretended CoT & \bf 74.41 \\
    & Knowledge Enhancement & 71.72 \\
    \midrule
    \multirow{3}{*}{OPT$-{\rm 2.7b}$} 
    & PromptEOL & 69.30 \\
    & Pretended CoT & \bf 72.77 \\
    & Knowledge Enhancement & 63.83 \\
    \midrule
    \multirow{3}{*}{OPT$-{\rm 6.7b}$} 
    & PromptEOL & 72.10 \\
    & Pretended CoT & \bf 74.91 \\
    & Knowledge Enhancement & 74.60 \\
    \midrule
    \multirow{3}{*}{OPT$-{\rm 13b}$} 
    & PromptEOL & 71.86 \\
    & Pretended CoT & 75.44 \\
    & Knowledge Enhancement & \bf 75.58 \\
    \bottomrule
\end{tabular}
\label{tab:opt_sts}
\end{table}

For PLMs of all sizes, optimal performance is invariably attained through either Pretended CoT or Knowledge Enhancement, reaffirming the scalability and adaptability of our strategies. Given that no adjustments are made to the manual templates as outlined in Table~\ref{tab:templates} based on model size disparities, this performance uplift showcases the exceptional plug-and-play nature of Pretended CoT and Knowledge Enhancement, underscoring their potential for seamless integration into new application scenarios with minimal additional effort.

Furthermore, the experimental outcomes elucidate that Pretended CoT consistently delivers a more stable enhancement to the models' raw embeddings compared to Knowledge Enhancement. It exhibits a universally positive impact across all PLMs evaluated in this study. On the contrary, Knowledge Enhancement does not uniformly surpass the baseline established by PromptEOL, particularly in smaller-scale models such as OPT$_{\rm 350m}$. However, as model size increases to 13 billion parameters, Knowledge Enhancement demonstrates superior efficacy over Pretended CoT. This trend suggests that Knowledge Enhancement, which encompasses a broader token range and a more complex vocabulary, necessitates a higher level of textual comprehension from the model to effectively inject prior information, thus benefiting more from increased model capacity.

\section{Analysis}

This chapter delves into the underlying factors contributing to the success of Pretended CoT and Knowledge Enhancement strategies introduced in this study. We initiate our analysis by exploring whether these methodologies yield sentence embeddings with enhanced distribution characteristics. This examination, predicated on alignment and uniformity metrics that measure the semantic space properties of PLMs, is detailed in Section~\ref{sec:align_uniform}. Subsequently, in Section~\ref{sec:attention}, we employ a randomly selected sample to visualize the attention allocation during sentence representation computation, aiming to ascertain the effectiveness of Knowledge Enhancement in directing the model's focus towards critical elements of the original sentence.

\subsection{Alignment and Uniformity}
\label{sec:align_uniform}

In the domain of sentence representation, alignment and uniformity are pivotal metrics for appraising the quality of embeddings. For a given data distribution $p_{data}$, alignment measures the proximity of semantically similar samples by calculating the expected squared distance between two akin sentences ($x$ and $x^+$) mapped to the embedding space by encoder $f$, as illustrated below:
\begin{equation}
\label{eq:align}
    \ell_{align}\triangleq \underset{(x, x^+)\sim p_{data}}{\mathbb{E}} \Vert f(x) - f(x^+) \Vert^2
\end{equation}

Uniformity, in contrast, focuses on the distribution's evenness across the semantic space, encouraging unrelated embeddings to be as distant from each other as possible, with its formula defined as:
\begin{equation}
\label{eq:uniform}
    \ell_{uniform}\triangleq{\log} \underset{~~~x, y\stackrel{i.i.d.}{\sim} p_{data}}{\mathbb{E}}   e^{-2\Vert f(x)-f(y) \Vert^2}
\end{equation}

With LLaMA2$_{\rm 7b}$ as the backbone, we evaluate the average Spearman correlation scores alongside alignment and uniformity metrics for PromptEOL, Pretended CoT, and Knowledge Enhancement. These assessments involve utilizing the output vectors from either the PLM's last or penultimate hidden layers as embeddings. The results are documented in Table~\ref{tab:align_uniform}.
\begin{table}[htbp]
\caption{Comparative analysis of different sentence representation methods on LLaMA2$_{\rm 7b}$. For Spearman, higher values indicate better performance, whereas for Alignment and Uniformity, lower values are desirable.}
\renewcommand
\arraystretch{1.2}
\centering
\setlength{\tabcolsep}{2pt}
\begin{adjustbox}{width=1.0\linewidth, center}
\begin{tabular}{ccccc}
\hline
\bf Methods & \bf Layer Index & \bf Spearman & \bf Alignment & \bf Uniformity \\
\hline
PromptEOL & -1 & 71.66 & 0.0557 & -0.9302 \\
Pretended CoT & -1 & 72.73 & 0.0533 & -0.9555 \\
Knowledge Enhancement & -1 & \bf 77.15 & \bf 0.0528 & \bf -0.9996 \\
\hline
PromptEOL & -2 & 76.30 & 0.0933 & -1.4888 \\
Pretended CoT & -2 & 78.95 & 0.0919 & -1.5817 \\
Knowledge Enhancement & -2 & \bf 80.34 & \bf 0.0908 & \bf -1.6742 \\
\hline
\end{tabular}
\end{adjustbox}
\label{tab:align_uniform}
\end{table}

The ``Layer Index" column in Table~\ref{tab:align_uniform} specifies the selected hidden layer for extracting the PLM's sentence representations, with -1 denoting the last layer and -2 referring to the penultimate layer. The ``Spearman" column retains its meaning from previous sections, signifying the average performance across seven STS benchmarks for the stated methods. Alignment and uniformity are computed via the STS-B test set, which contains 1,379 sentences pairs with similarity scores ranging from 0 to 5. For uniformity calculation, the entire dataset is utilized, while for alignment, a threshold of 4.5 is applied to select semantically comparable sample pairs.

The experimental data reveal that Pretended CoT and Knowledge Enhancement, both of which outperform PromptEOL in Spearman correlation, also exhibit superior results in alignment and uniformity metrics. Notably, the Spearman, Alignment, and Uniformity metrics collectively manifest a direct positive correlation, underscoring that our proposed strategies effectively refine the semantic space properties of sentence embeddings, thereby elevating their overall quality.

\subsection{Attention Allocation}
\label{sec:attention}

In Section~\ref{sec:prompt}, we have showcased the specific form of our Knowledge Enhancement strategy and expounded on its design rationale. Our objective is to fully harness the contextual learning and instruction following capabilities of generative PLMs by infusing human text summarization expertise directly into the manual template, thereby strengthening the model's focus on key semantic units.

\begin{figure*}[ht]
\centering \includegraphics[width=1.0\linewidth]{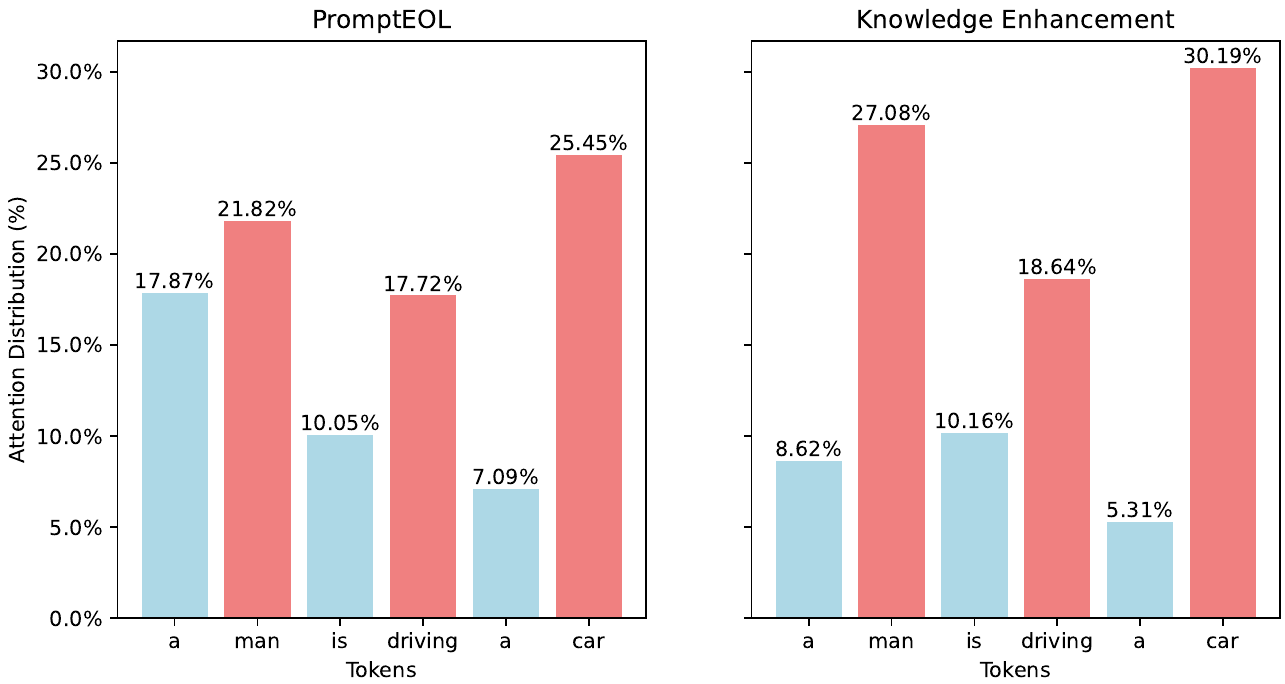}
\caption{Distribution of each token's contribution to sentence embeddings derived from PromptEOL and Knowledge Enhancement.}
\label{fig:bar}
\end{figure*}

The compelling empirical evidence presented in Section~\ref{sec:experiment} has provided robust validation for the efficacy of this approach. To facilitate a more transparent and intuitive grasp of the working principle behind Knowledge Enhancement, we conduct an analysis on a randomly selected sentence (``a man is driving a car") from the STS-B test set. Employing LLaMA2$_{\rm 7b}$ as the base model, we compute the cosine similarity between the sentence embeddings produced by PromptEOL and Knowledge Enhancement, which are output vectors corresponding to the prompt's final token, and the output vectors of each token in the original sentence. According to the self-attention computation mechanism of the Transformer \cite{Transformer-NIPS-2017} architecture $\text{softmax}(\frac{QK^T}{\sqrt d_k}V)$, a higher similarity value indicates more focused attention on a given token.

We classify tokens into two categories based on their semantic significance and sequentially tally the proportion of each token's similarity score to the total. Figure~\ref{fig:bar} visualizes these results, where red bars represent tokens that form the subject, predicate, and object of the original sentence—elements fundamental to its semantics integrity. Blue bars, on the other hand, denote tokens that primarily functioning as modifiers, which typically carry less semantic weight. Evidently, the Knowledge Enhancement statistics demonstrate a higher proportion allocated to red bars compared to PromptEOL. This discrepancy signifies that Knowledge Enhancement, when formulating sentence representation, assigns more weight to the sentence's core lexicon, cohering well with our initial design intentions.

\section{Conclusion}
\label{sec:conclusion}

In this study, we initially elucidate through a series of experiments that discussions surrounding the Explicit One-word Limitation should concentrate on the direct inference scenarios of generative Pre-trained Language Models (PLMs), as this technique is not inherently critical for discriminative models or for the fine-tuning of generative PLMs. Building on this foundation, we propose two innovative prompt engineering strategies to further enhance the quality of generative PLMs' raw embeddings: Pretended Chain of Thought and Knowledge Enhancement. Our extensive experimentation confirms the efficacy and universality of these methods across a spectrum of models with varying types and parameter sizes. Remarkably, when paired with 7B-scale PLMs, our approach not only achieve performance on par with or even superior to unsupervised contrastive learning methods that have undergone fine-tuning but also demonstrate a reduction in GPU memory consumption. Considering the plug-and-play characteristics of Pretended Chain of Thought and Knowledge Enhancement, they possess a broad application prospect. Additionally, through a detailed examination of alignment and uniformity metrics alongside attention distribution analyses, we illuminate the contributory factors underpinning the success of these two schemes. Our code has been open-sourced for researchers in the related fields to verify and conduct further experiments.

%
%
%
%

\end{document}